\documentclass[a4paper, 10pt, conference]{ieeeconf}      

\IEEEoverridecommandlockouts                              
 \usepackage{xcolor}  
 \usepackage[onehalfspacing]{setspace}
 \usepackage{float}
 \usepackage{graphicx}
 \usepackage{caption}
 \usepackage{subfig}

\title{\LARGE \bf
Performance of Image Registration Tools on High-Resolution 3D Brain Images
}

\author{Abdullah Nazib\,$^{1}$, James Galloway\,$^{1}$, Clinton Fookes\,$^{1}$, Dimitri Perrin\,$^{1,*}$
\thanks{*This work was  supported by Queensland University of Technology}
\thanks{$^{1}$ Faculty of Electrical Engineering and Computer Science,
	Queensland University of Technology. 
        {\tt\small dimitri.perrin@qut.edu.au}}%
}

\begin{document}

\pagenumbering{1}
\maketitle
\thispagestyle{empty}
\pagestyle{empty}

\begin{abstract}
Recent progress in tissue clearing has allowed for the imaging of entire organs at single-cell resolution. These methods produce very large 3D images (several gigabytes for a whole mouse brain). A necessary step in analysing these images is registration across samples. Existing methods of registration were developed for lower resolution image modalities (e.g. MRI) and it is unclear whether their performance and accuracy is satisfactory at this larger scale. In this study, we used data from different mouse brains cleared with the CUBIC protocol to evaluate five freely available image registration tools. We used several performance metrics to assess accuracy, and completion time as a measure of efficiency. The results of this evaluation suggest that the ANTS registration tool provides the best registration accuracy while Elastix has the highest computational efficiency among the methods with an acceptable accuracy. The results also highlight the need to develop new registration methods optimised for these high-resolution 3D images.
\end{abstract}

\section{INTRODUCTION}


Tissue clearing methods provide system-level identification and analysis of cellular circuits in biological samples maintaining structural integrity. The images obtained with these new methods are several orders of magnitude larger than those acquired using standard methods. A typical MRI image, e.g from the LPBA40 dataset \cite{Shattuck2008}, has a voxel resolution of $0.86\times0.86\times1.5$ mm$^3$, for a total of approximately 8 million voxels. The dataset used in this study is known as CUBIC dataset\cite{Susaki2014}. Samples in this dataset have a voxel resolution of $6.45\times6.45\times10$ $\mu$m$^3$, for a total of more than 3 billion voxels \cite{Susaki2014}. A single brain sample is over 6 gigabytes. This creates significant challenges for the downstream analysis of these images.

In medical image analysis, image registration is a commonly used process that transform different data sample into a common co-ordinate system and enables direct comparison across the samples. Registration is a computationally expensive iterative process requiring similarity measurment and transformation calculation in each optimization step. State-of-the-art image registration tools pre-date large scale tissue cleared dataset. Therefore it is important to assess their suitability for high-resolution tissue cleared data.


The objective of this paper is to evaluate such image registration tools on CUBIC dataset. We focus on five well-known and freely available tools: Advanced Normalization Tools (ANTS), Insight Registration Toolkit (IRTK), Automated Image Registration (AIR), Elastix and NiftyReg. The performance of the registration tools are evaluated in terms of accuracy (both quantitatively and qualitatively) and computational efficiency \cite{Alterovitz2006}.

\section{Materials and methods}

\subsection{CUBIC Dataset}

For this evaluation, the CUBIC dataset is downscaled into 25\% resolution since the computational cost of registering 100\% is prohibitively large. In addition to the 25\%-resolution files, we also created files at 10\%, 15\%, and 20\% resolution. We also created 100\%-resolution files for the D-V(Dorsal to Ventral) stacks for further analysis of the top-performing tools.
\subsection{Artificial Data}
To complement the assessment performed on intersubject data, we also created an artificial dataset by deforming the files generated above (so that each `deformed brain' can then be registered back to its original version to allow for an accurate validation against a known ground truth deformation). The deformed files were generated using simple Gaussian deformation. A Gaussian distributed deformation vector field was generated from a pixel grid. Pixel intensities were then interpolated over the Gaussian distributed deformation field. Repeatedly applying the Gaussian deformation field over all slices in a given 3D volume generated a Gaussian-deformed volume which was used for evaluation.

\subsection{Registration Process and Settings}
The same process was used for all five registration tools, both contexts (intersubject registration, and intrasubject registration of the artificially-deformed data), and all file sizes: rigid registration, followed by affine registration and, lastly, deformable registration. In the intersubject context, one of the brains was used as reference and the other two were aligned to it. In the intrasubject context, the original samples were used as reference and the deformed ones were aligned back to those. The tests for both contexts were performed on a Core i7 3.4 GHz, 16 GB RAM workstation with 64-bit Ubuntu 16.04 OS, as well as on an HPC cluster. Because the registration tools are not parallelised, we used a single compute core (with 2.66GHz 64bit Intel Xeon processors and 256GB memory).        

The performance of the tools used in this evaluation can be affected by parameter choice, and we have therefore considered this carefully. For IRTK, we followed the same settings as \cite{xu2016}, except B-spline control points. The control point spacing is set to 5mm, which is the highest possible value for this method.

For the AIR \cite{Woods1992},\cite{Woods1998} parameter settings, we applied the default threshold values for rigid and affine transformation.
Using affine registered parameters as the starting points for non-linear registration, second order and third order non-linear registrations are performed consequtively.

Elastix \cite{Klein2010} includes a number of transformation models (rigid, affine with different degrees of freedom, B-spline with physics based control points in uniform and non-uniform grids), as well as a several optimisation methods. In this evaluation, we used the same parameter settings as \cite{Hammelrath2016} for rigid, affine and non-linear registration.

The ANTS \cite{Avants2008} parameters are derived from example scripts. We used cross-correlation as a similarity measure, resolution level 3, and 100 iterations for each sampling level. 

As for IRTK \cite{Rueckert1999}, the parameter choice for NiftyReg \cite{Modat2010} is based on \cite{xu2016}. The number of iterations was set at 1000 for free-form deformation and we used an intensity threshold of 500 for both source and target image.
\subsection{Evaluation Measures}
There are multiple methods of measuring similarity between images. In this study, we have used cross correlation and the Mutual Information. 

The mutual information is a widely accepted method of measuring information contents in two comparing distribution. It can also be applied in registration to measure the similarity between a target and reference brain. In medical image registration, MI is extensively used to compare images of different modalities. Here, because the cell signal is already discrete in nature and their numbers are different in comparing brains, we used MI value directly to measure similar information between them. Mathematically mutual information is:
\begin{equation}
	I(X,Y)=\sum_{y \in Y}{\sum_{x \in X}{p(x,y).\log{\frac{p(x,y)}{p(x).p(y)}}}}
\end{equation}


Cross-correlation is a standard method of estimating the degree of correlation between two different entities.
It is used extensively in signal processing to correlate signal properties:
\begin{equation}
R_c= \frac{\sum{(X-\mu _x)*(Y-\mu _y)}}{\sqrt{\sum{(X-\mu _x)}^2}*\sqrt{\sum{(Y-\mu _y)}^2}}
\end{equation}

To complement the quantitative information obtained from these two metrics, we also qualitatively assessed the results through visual inspection of overlaid registered and reference images.

To assess registration efficiency, we also measured the computation time for each method and each input size.   
\section{Results} 
\subsection{Quantitative Registration Accuracy}
\subsubsection{Inter-Brain Registration}
For the inter-brain registration program, brain sample 3 was used as the reference brain, and brain samples 1 and 2 were aligned to this reference. 

Tables \ref{InterCross} show the averaged cross-correlation and mutual information measures of each registration tool.
Table \ref{InterCross} show that ANTS obtained the best cross-correlation average, at 0.93 for brain 1 and 0.95 for brain 2. Elastix also performed well, at 0.92 for brain 1 and 0.94 for brain 2. The average cross-correlation for NiftyReg was under 0.8, while both IRTK and AIR scored under 0.5. 
Table \ref{InterCross} also shows that Elastix obtained the best mutual information score, at 2.41 for brain 1 and 2.28 for brain 2. Both ANTS and NiftyReg also scored between 1.3 and 1.5, while the scores for IRTK and AIR remained under 1. 

At 100\%-resolution, we tested Elastix and ANTS on an HPC system, running a job with 256Gb and 100-hour wall time. We used 3D images from the D-V stacks for brains 1 and 2, and registered them to brain 3. We then repeated the process for the V-D images. The Elastix results are shown in Table \ref{Inter100}, and are similar to those observed at 25\%-resolution. ANTS failed to register any brain, even after extended the wall time to 200 hours.

\begin{table}[htb]
\begin{center}
	\tiny
		\caption{\small Mean CC and MI for inter-brain pipeline}
		\label{InterCross}
		\begin{tabular}{l c c c c}
			\hline
			Methods & brain 1(CC) &brain 1(MI) & brain 2(CC) & brain 2(MI)\\
			\hline
			ANTS	&0.9303		&1.4317			&0.9546		&1.5189\\
			AIR		&0.0124		&0.8879			&0.0154		&0.9172\\
			Elastix	&0.9163		&2.4115			&0.9388		&2.2887\\
			NiftyReg&0.7487		&1.4625			&0.7645		&1.3856\\
			IRTK	&0.3268		&0.4563			&0.3484		&0.3456\\
			\hline
		\end{tabular}
\end{center}
\end{table}

\subsubsection{Intra-Brain Registration}
Since there is no absolute ground truth when registering two images from distinct brain samples, the evaluation of inter-brain registration only capture part of the overall performance. To remedy this, we have performed intra-brain registration, where each deformed brain sample is registered to its original version. 

ANTS remained the best performer in terms of cross-correlation, and Elastix continued to outperform ANTS in all but the 10\% resolution when considering the Dice similarity coefficient. Finally, Elastix scored highly across both metrics and all resolutions. 
Tables \ref{IntraCross} shows the average cross-correlation and Mutual Information results, and confirm the overall trend of these results.

\begin{table}[htb]
	\tiny

	\begin{center}
		\caption{\small Mean CC and MI for intra-brain pipeline}
		\label{IntraCross}
		\begin{tabular}{l c c c c c c}
			\hline
			Methods & brain 1(CC) & brain1(MI) &Brain 2(CC) 	&brain 2(MI) & brain 3(CC) 	&brain 3(MI)\\
			\hline
			ANTS	&0.9805			&1.4136			&0.9822		&1.9872			&0.9823		&1.8591\\
			AIR		&0.1657			&1.2865			&0.1651		&1.2340			&0.0097		&1.2451\\
			Elastix	&0.9719			&2.2385			&0.9692		&2.3453			&0.9713		&2.7193\\
			NiftyReg&0.9411			&1.4952			&0.9494		&1.8642			&0.9394		&2.4223\\
			IRTK	&0.5156			&0.8534			&0.5160		&0.5156			&0.4881		&0.7286\\
			\hline
		\end{tabular}
	\end{center}

\end{table}

\subsection{Visual Analysis}
\begin{figure}[h]
	\centering\includegraphics[width=6cm]{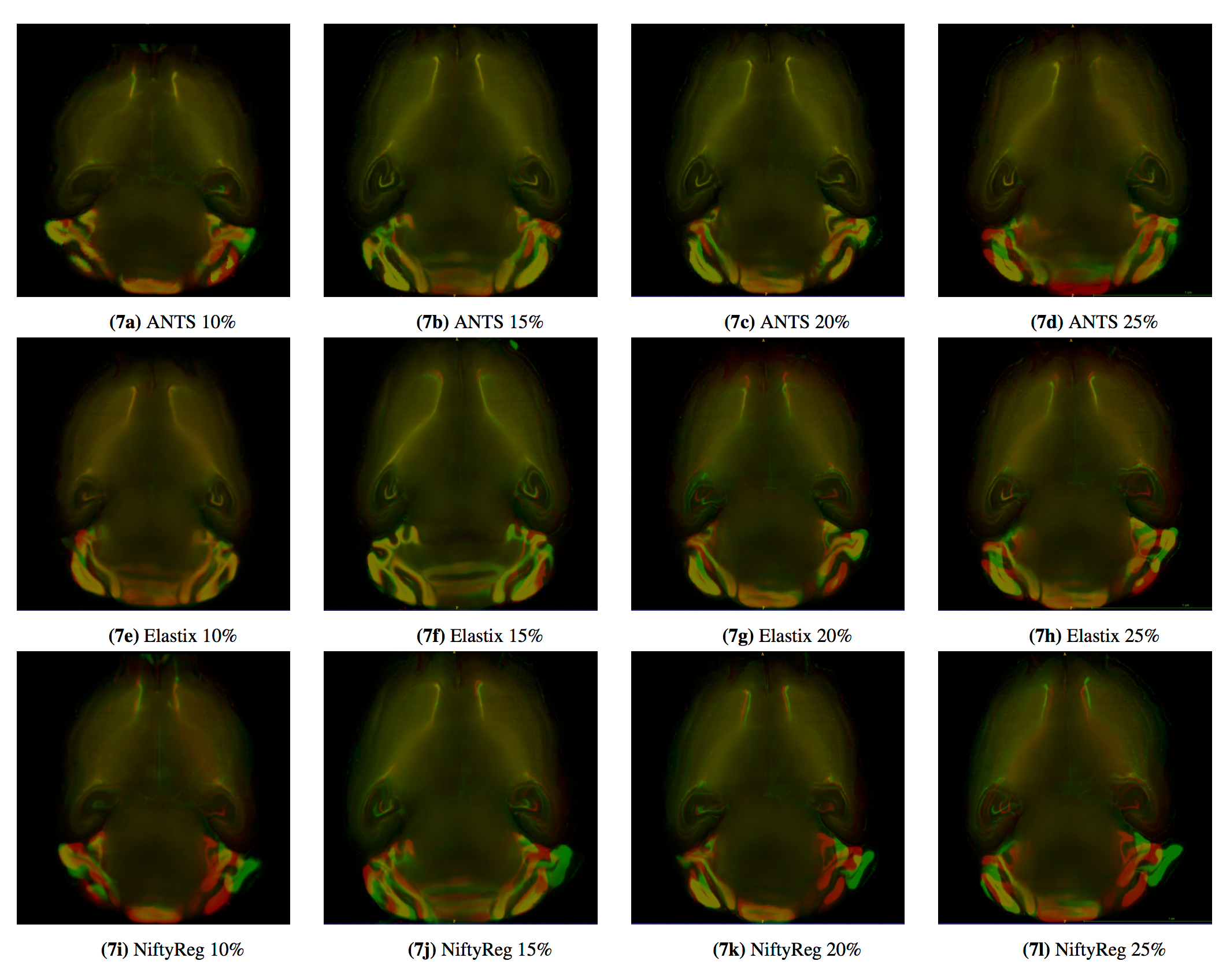}
	\caption{Qualitative results for inter-brain registration.}
	\label{fig:InterVis}	
\end{figure}

\begin{figure}[h]
	\centering\includegraphics[width=6cm]{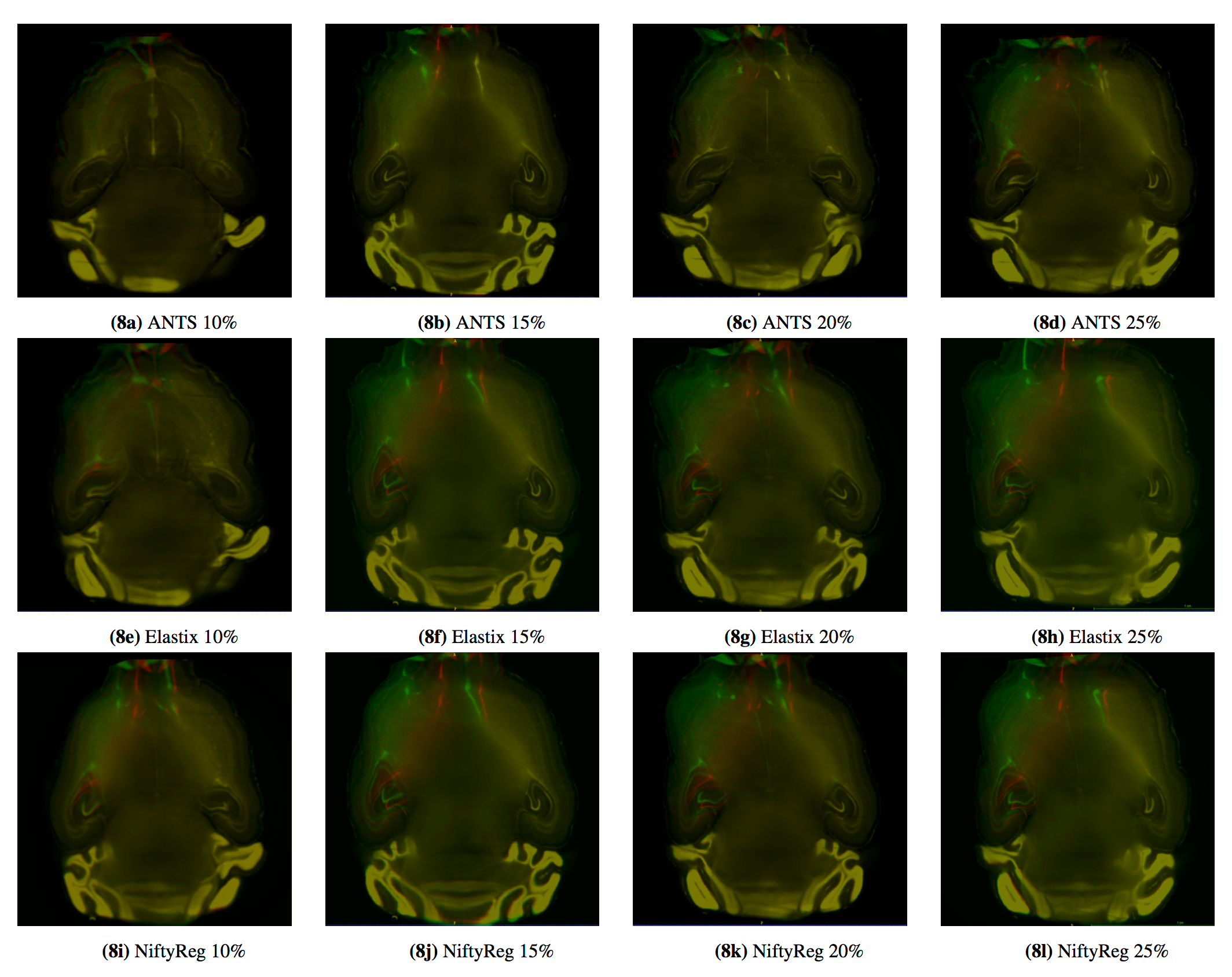}
	\caption{Qualitative results for intra-brain registration.}
	\label{fig:IntraVis}
\end{figure}

Our quantitative results are complemented with a qualitative inspection of the alignments.
Figures \ref{fig:InterVis} and \ref{fig:IntraVis} are visual representations of the results for the three top-performing methods (ANTS, Elastix, and NiftyReg), where the reference brains (red) are overlaid with their corresponding aligned brains (green). 

Figure \ref{fig:InterVis} shows the inter-brain registration results.  The first row shows all four resolutions registered by ANTS (Figure \ref{fig:InterVis}a--d). There are visible differences in the cerebellum, but this is expected for all methods and is not a problem for subsequent analyses, which typically focussed on other regions. There are also smaller differences in other regions such as the hippocampal formation, and in particular the dentate gyrus.

Elastix gave good results at 10\% resolution, but as the resolution increased it underperformed against ANTS. The mismatch of the hippocampal formation is more pronounced, and other regions towards the cerebral cortex are also misaligned. The same issues are present in the NiftyReg results.

Similar results can be observed for the intra-brain context (Figure \ref{fig:IntraVis}). The deformation is larger than the typical difference between two mouse brains, but ANTS still performs relatively well, and visibly better than Elastix and NiftyReg.

Overall, taking together the quantitative and qualitative results, ANTS is the best performer in this evaluation.  


\subsection{Computational Efficiency}
The computational efficiency of each registration tool is shown in Tables \ref{InterPerformHPC} and \ref{IntraPerformHPC}. 
We performed experiments using HPC jobs with 32GB memory for 10\%- and 15\%-resolution files, and 64GB memory for 20\%- and 25\%-resolution files.
We did not observe any direct relationship between accuracy and efficiency. AIR is very efficient but, not very accurate on this type of data. IRTK had similar accuracy issues, but also was inefficient, with a 25\% resolution dataset taking more than a day to process.

Amongst the top three performers in terms of accuracy, ANTS is the least efficient, with completion times of more than 12 hours for intra-brain registration and more than 8 hours for inter-brain registration at 25\% resolution. 
Elastix proved to be the most efficient of the three, with all resolution sizes being processed in less than an hour on HPC environment.

To confirm that these results are not simply a memory bottleneck, we ran pipelines with all resolution in a workstation with 16GB memory and obtain similar efficiency pattern. Similar pattern ensures that simply adding more memory does not significantly improve performance, and confirms bottleneck is computational.

\begin{table}[h]
	\tiny
	\begin{center}
		\caption{Computation time (inter-brain pipeline, HPC)}
		\label{InterPerformHPC}
		\begin{tabular}{l c c c c}
			\hline
			Methods & 10\% & 15\% & 20\% & 25\%\\
			\hline
			ANTS		&00:42:45   &01:38:31   &4:56:48	&08:40:11\\
			AIR			&00:00:27   &00:06:52  	&00:05:04   &00:37:15\\
			Elastix 	&00:02:17   &00:05:15 	&00:08:36	&00:16:21\\ 
			NiftyReg	&00:04:35   &00:19:40 	&01:39:20   &01:45:50\\
			IRTK		&00:03:52   &00:26:57 	&02:28:10 	&25:42:26\\
			\hline
		\end{tabular}
	\end{center}
\end{table}

\begin{table}[h]
	\tiny
	\begin{center}
		\caption{Computation time (intra-brain pipeline, HPC)}
		\label{IntraPerformHPC}
		\begin{tabular}{l c c c c}
			\hline
			Methods & 10\% & 15\% & 20\% & 25\%\\
			\hline
			ANTS		&01:23:44   &04:13:18   &08:24:49	&12:24:28\\
			AIR			&00:00:31   &00:25:51 	&00:10:18  	&00:50:14\\
			Elastix 	&00:05:03  	&00:06:30 	&00:17:42	&00:24:30\\ 
			NiftyReg	&00:16:32   &00:44:55	&02:12:56  	&04:06:51\\
			IRTK		&00:02:42   &03:09:26 	&06:05:20	&30:22:47\\
			\hline
		\end{tabular}
	\end{center}
\end{table}

\subsection{Parameter exploration}

To confirm that the choice of parameters did not affect our results, we also performed a parameter exploration for the most promising two tools, ANTS and Elastix, on the 25\%-resolution files. All tests were performed on 32GB memory, 100-hour wall time HPC jobs, with larger resources used if initial tests failed.

The additional configurations and corresponding results for ANTS results are shown in Table \ref{Table:ANTS_param_opt}. For our data, these parameter sets have a limited impact on the registration quality: ANTS performed well under all configurations. Increasing the number of optimisation levels, and using mutual information as the similarity measure, improved the quality of the results. 
Mutual information was also associated with lower computation time. Inspection of the logs revealed that this is because the registration process converged faster. Cross-correlation takes longer, and there is one configuration where ANTS was not able to finish the registration within the 200-hour wall time. 

Similar experiments on Elastix are shown in Table \ref{Table:Elastix_param_opt}. Unlike ANTS, impact of parameter choice on the quality of results are not observed in Elastix. 
The computation time remained largely similar, and the quantitative results were better with cross-correlation as the similarity measure. As one would expect, quantitative results were improved by increasing the number of optimisation levels, up to the point where Elastix could not complete the registration. However, visual inspection of the results, 
reveal that deformations induced by these altered parameters were too drastic, and created anatomically incorrect features. The initial parameter set was the best configuration.

\begin{table}[htb]
	\tiny
	\begin{center}
		\caption{ANTS Parameter Optimization}
		\label{Table:ANTS_param_opt}
		\begin{tabular}{l c c c c c c c}
			\hline		
			& Parameter Set &Similarity 	&Interp 	& Levels 	& Cross 	& MI 	& Time\\ 	
			\hline
			&1			&CC,1,2			&Linear		&1,1,3		&0.7070		&0.6702	&03:59:53\\
			&2			&MI,1,32		&Bspline	&3,3,3		&0.7868		&0.6208	&01:04:58\\
			&3			&MI,1,64		&Bspline	&5,5,5		&0.7359		&0.8623	&01:16:06\\
			&4			&CC,1,5			&Bspline	&5,5,5		&\multicolumn{3}{c}{Unable to complete within 200h.}\\
			&5			&CC,1,2			&Bspline	&3,3,5		&0.6868		&0.6557	&38:59:59\\
			&6			&MI,1,32		&Bspline	&3,3,5		&0.7429		&0.8622	&01:02:36\\
			&7			&MI,1,32		&Linear		&3,3,3		&0.7825		&0.9257	&01:00:25\\
			\hline
		\end{tabular}
	\end{center}
\end{table}

\begin{table}[htb]
	\tiny
	\begin{center}
		\caption{Elastix Parameter Optimization}
		\label{Table:Elastix_param_opt}
		\begin{tabular}{l c c c c c c c}
			\hline		
			& Parameter Set &Similarity 	&Iterations 	& Levels 	& Cross 	& MI 	& Time\\ 	
			\hline
			&1			&CC				&1000		&3			&0.9301		&0.8854	&01:54:04\\
			&2			&MI,32			&1000		&3			&0.8752		&1.0234	&02:03:46\\
			&3			&MI,64			&1000		&5			&0.8695		&1.1337	&02:41:27\\
			&4			&CC				&1000		&5			&0.9538		&0.9933	&02:41:08\\
			&5			&CC				&2500		&6			&\multicolumn{3}{c}{Unable to complete. Program stops.}\\
			&6			&MI,32			&2500		&6			&\multicolumn{3}{c}{Unable to complete. Program stops.}\\		
			\hline
		\end{tabular}
	\end{center}
\end{table}

\begin{table}[htb]
	\tiny
	\begin{center}
		\caption{Elastix results at 100\%-resolution (inter-brain pipeline)}
		\label{Inter100}
		\begin{tabular}{l c c c c}
			\hline
			Samples & Cross-correlation &Mutual information &Time\\
			\hline
			Brain 1 D-V	&0.8412 		&0.8018				&27:16:05	\\
			Brain 2 D-V	&0.8676 		&0.8385				&27:42:14\\
			Brain 1 V-D	&0.4403			&0.4781				&32:39:42\\
			Brain 2 V-D &0.4278			&0.4852				&32:39:42\\ 
			\hline
		\end{tabular}
	\end{center}
\end{table}

\section{Conclusion}

In this study, we analysed five registration tools on high-resolution whole-brain images obtained with the CUBIC method. Three of these (ANTS, Elastix and NiftyReg) gave good quantitative results. IRTK proved to be unstable producing completely black images in some case and over-saturated signals in some other case. AIR proved unsuitable for discrete CUBIC dataset since it registers images by minimizing voxel ratio.
Overall, taking together all the evaluation measures, ANTS appears to be the best choice for the registration of 3D samples obtained with tissue clearing and LSFM imaging when the overall analysis is not degraded by registering at 25\% resolution, as in \cite{Tatsuki2016}. Using it with mutual information as the similarity measure gave the best computational performance. In datasets where registration at 100\% resolution is required, it may be substituted with Elastix. However, even Elastix struggles with these larger files. It took more than a day per registration, which limits its practical use. This study therefore highlights the need to develop dedicated tools for this new type of data. New approaches such as deep learning are being considered in an image registration context \cite{Wu2016}, and may provide suitable alternatives.

\bibliographystyle{ieeeconf}
\bibliography{new_refs.bib}
 
\end{document}